\documentclass{article}
\PassOptionsToPackage{compress,sort,authoryear}{natbib}

\newcommand{\trainx}{x_{\text{tr}}}
\newcommand{\trainy}{y_{\text{tr}}}

\newcommand{\trainz}{z_{\text{tr}}}
\newcommand{\testz}{z_{\text{te}}}

\newcommand{\testx}{x_{\text{te}}}
\newcommand{\testy}{y_{\text{te}}}

\newcommand{\testpredy}{\hat{y}_{\text{te}}}

\newcommand{\infl}{\mathrm{Infl}}
\newcommand{\deltainflij}[2]{\Delta\mathrm{Infl}({#1},{#2})}

\newcommand{\br}[1]{\left(#1\right)}

\usepackage[cm]{fullpage}

\usepackage[final]{neurips_2024}

\usepackage{amsmath}
\usepackage{mathtools}
\usepackage[utf8]{inputenc} % allow utf-8 input
\usepackage[T1]{fontenc}    % use 8-bit T1 fonts
\usepackage{url}            % simple URL typesetting
\usepackage{booktabs}       % professional-quality tables
\usepackage{amsfonts}       % blackboard math symbols
\usepackage{nicefrac}       % compact symbols for 1/2, etc.
\usepackage{microtype}      % microtypography
\usepackage{xcolor}         % colors

\usepackage[pagebackref]{hyperref}
\usepackage[authoryear]{natbib}
\usepackage{url}
\usepackage{booktabs}
\usepackage{makecell}
\usepackage{algorithm}      % algo
\usepackage{algpseudocode}  % algo
\usepackage{graphicx}
\usepackage{color, colortbl}
\usepackage[dvipsnames]{xcolor}
\usepackage{xspace}
\usepackage{pifont}
\usepackage{enumitem}
\usepackage{cleveref}
\usepackage{float}  % table
\usepackage{multirow}  % table
\definecolor{softgreen}{rgb}{0.13, 0.55, 0.13}
\definecolor{softred}{rgb}{0.8, 0.13, 0.13}
\newcommand{\delinfl}{\(\Delta-\mathrm{Influence}\)\xspace}
\definecolor{kleinblue}{RGB}{0, 47, 167} 
\definecolor{kleinblue2}{RGB}{20, 20, 125} 
\definecolor{kleinred}{HTML}{bc1919}

\title{\Large \(\Delta\)-Influence: Unlearning Poisons via Influence Functions}

\author{
  Wenjie Li \\
  ShanghaiTech University \\
  \texttt{} \\
  \And
  Jiawei Li \\
  Tsinghua University \\
  \And
  Christian Schroeder de Witt \\
  University of Oxford \\
  \texttt{} \\
  \And
  Pengcheng Zeng \\
  ShanghaiTech University \\
  \texttt{} \\
  \And
  Ameya Prabhu \\
  T\"ubingen AI Center, 
  University of T\"ubingen \\
  \texttt{} \\
  \And
  Amartya Sanyal \\
  University of Copenhagen \\
}

\begin{document}

\maketitle
\vspace{-0.4cm}
\begin{abstract}
\vspace{-0.2cm}
Addressing data integrity challenges, such as unlearning the effects of data poisoning after model training, is necessary for the reliable deployment of machine learning models. State-of-the-art influence functions, such as EK-FAC~\citep{grosse_studying_2023} and TRAK~\citep{park_2023_trak}, often fail to accurately attribute abnormal model behavior to specific poisoned training data responsible for the data poisoning attack. In addition, traditional unlearning algorithms often struggle to effectively remove the influence of poisoned samples~\citep{pawelczyk_machine_2024}, particularly when only a few affected examples can be identified~\citep{goel_corrective_2024}. 
To address these challenges, we introduce $\Delta$-Influence, a novel approach that leverages influence functions to trace abnormal model behavior back to the responsible poisoned training data using just \textit{one} poisoned test example, without assuming any prior knowledge of the attack. $\Delta$-Influence applies data transformations that sever the link between poisoned training data and compromised test points without significantly affecting clean data. This allows detecting large negative shifts in influence scores following data transformations, a phenomenon we term as influence collapse,  thereby accurately identifying poisoned training data. Unlearning this subset, \textit{e.g.} through retraining, effectively eliminates the data poisoning. We validate our method across three vision-based poisoning attacks and three datasets, benchmarking against five detection algorithms and five unlearning strategies. We show that $\Delta$-Influence consistently achieves the best unlearning across all settings, showing the promise of influence functions for corrective unlearning. Our code is publicly available at: \url{https://github.com/Ruby-a07/delta-influence}
\end{abstract}
\begin{figure}[h]
    \centering
    \vspace{-0.4cm}
    \includegraphics[width=\linewidth]{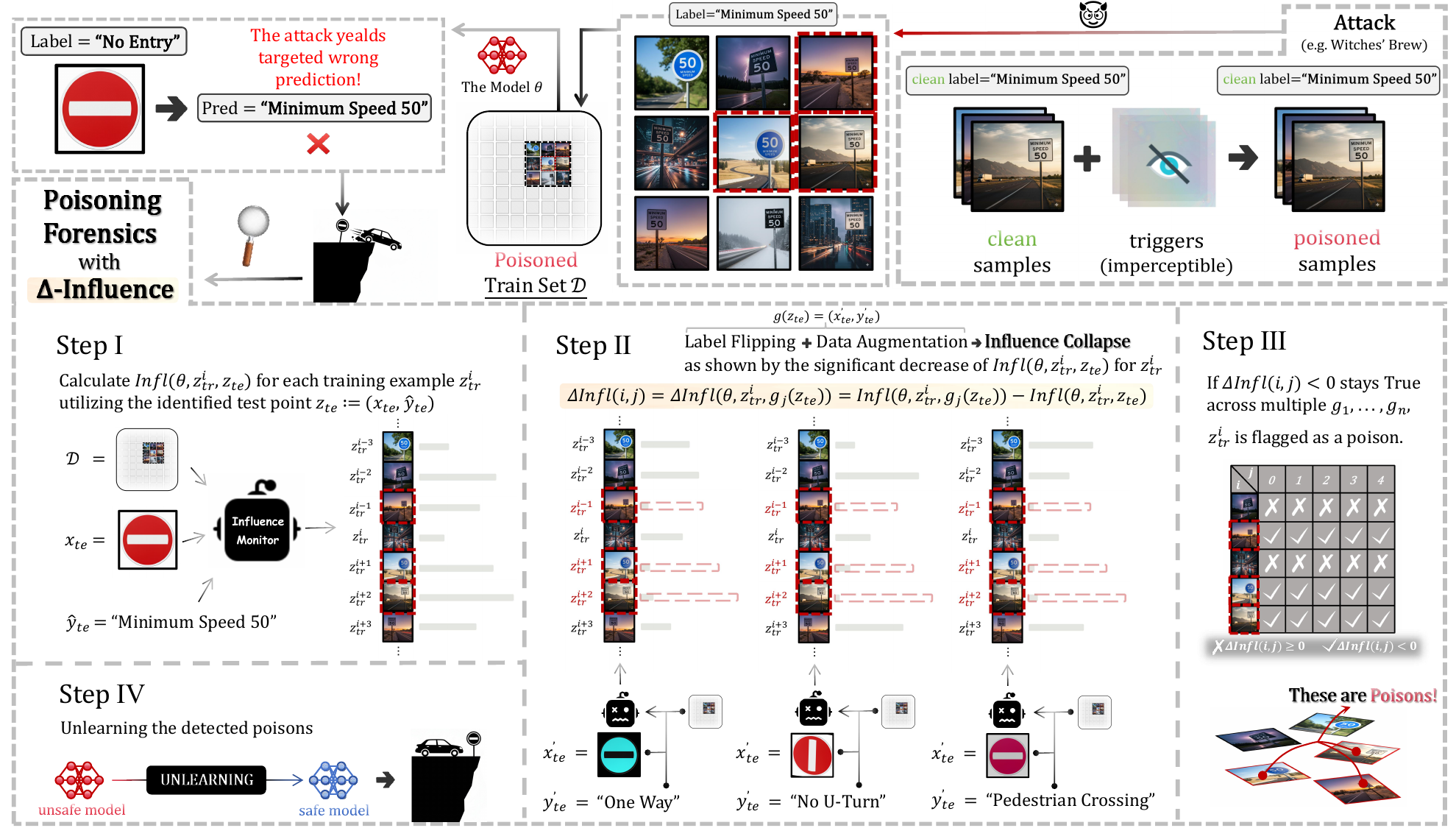}
    \vspace{-0.5cm}
    \caption{Given an affected test point, our goal is to identify the training points responsible for the poisoning, so that retraining without these points can remove the attack from the model. State-of-the-art methods like EK-FAC~\citep{grosse_studying_2023} detect only a few poisoned points with low precision, leaving the poisoning effect in the model and causing a large accuracy drop. Our method, \delinfl, outperforms existing approaches by successfully recovering the clean model without sacrificing accuracy.}
    \vspace{-0.4cm}
    \label{fig:framework}
\end{figure}
\section{Introduction}
\label{sec:intro}
\vspace{-0.15cm}

Machine learning models are increasingly deployed in critical sectors such as healthcare, finance, and autonomous systems~\citep{davenport_potential_2019, huang_deep_2020, chatila_trustworthy_2021, soori_artificial_2023}. This underscores the importance of ensuring model integrity and robustness against data poisoning attacks. In data poisoning, adversaries intentionally manipulate training data by introducing carefully crafted, often imperceptible modifications~\citep{chatila_trustworthy_2021}, leading to incorrect predictions or embedding specific malicious behaviors within the trained models~\citep{fan_survey_2022}. Given the large scale of modern datasets, identifying and removing all manipulated samples is typically impractical~\citep{nguyen_2024_towards, goel_corrective_2024}. Therefore, a viable approach involves detecting and attributing the impact of data poisoning to a small set of \emph{influential} training data points, which is \emph{unlearned} to mitigate the data poisoning attack.

The challenge of effective unlearning largely depends on the extent of knowledge about the data poisoning attack. For example, \citet{goel_corrective_2024} demonstrate that retraining a model after removing a randomly sampled subset containing half of the manipulated data fails to eliminate poisoning in relatively simple attacks like BadNet~\citep{gu_2017_badnets}. In contrast, retraining without the entire set of manipulated data successfully removes the attack. 
Furthermore, for more sophisticated poisoning strategies such as Witches' Brew~\citep{geiping_witches_2021}, \citet{pawelczyk_machine_2024} reveal that existing unlearning algorithms are ineffective unless the model is retrained without the full manipulated set, even when full access to the manipulated data is available.

Building upon the framework of \emph{Corrective Unlearning} introduced by \citet{goel_corrective_2024}, our work addresses the setting in which the defender has identified a small set of affected test points. We note that detecting such affected data is a practical trigger for realizing that unlearning is necessary and thereby initiating the unlearning process, which can be regarded as a form of \emph{poisoning forensics}: starting from a compromised output, we trace back to the culpable training examples whose removal neutralizes the attack.
In practice, such ``perpetrators'' typically surface through (i) deployment observation of anomalous behavior (\textit{e.g.}, a permission system granting administrative access to an unknown user, a stop sign being misclassified as a minimum speed-limit sign) or (ii) deliberate in-house stress testing (\textit{e.g.}, red‑teaming, white‑hat). 
A key advantage of \delinfl is that it requires only the logically unavoidable minimum that at least~\emph{one affected test point can be identified}; other methods, while effective in their respective settings, typically assume a larger identified set~\citep{Min_2025_ifrlhf, coalson_2025_ifguide}.
Leveraging this poisoned test point, our approach comprises two primary tasks: first, identifying a critical set of manipulated training points responsible for the compromised prediction; and second, applying unlearning algorithms to remove the influence of these points from the model.
% Additionally, the reason we target one affected test point rather than one poisoned training point, is that identifying poisoned training data can be substantially more difficult in complex clean-label attacks like Witches' Brew~\citep{geiping_witches_2021}. 
% Moreover, we empirically show that using one poisoned training point as the target can result in failed unlearning even when people have access to it (see Section~\ref{sec:training_or_test_target} for more details).
Our approach departs from prior unlearning works that often presuppose the availability of a ``forget set'', a subset of known poisoned training points~\citep{goel_2023_adversarial, kurmanji_2023_towards, foster_2023_fast}. We argue this assumption can be challenging to satisfy in practice, as identifying even one culpable training sample can be difficult, if not impossible, especially in complex clean-label attacks like Witches' Brew (as illustrated in the `Attack' panel of \Cref{fig:framework}).

Within this framework, \emph{influence functions}~\citep{koh_understanding_2017} serve as a natural tool for attributing model predictions to specific training data points. 
However, recent studies~\citep{grosse_studying_2023, nguyen_2024_bayesian, bae_2024_training} have indicated that state-of-the-art influence functions struggle to accurately identify the manipulated data when used in a naive manner. 
Our experiments in~\Cref{sec:exp} also corroborate this finding. 
To address this limitation, we introduce \delinfl, a novel approach that enhances influence functions to reliably identify a critical set of training data points necessary for unlearning data poisoning without compromising model accuracy. 
Instead of directly calculating each training point's influence on a poisoned test point,~\delinfl assesses the change in influence scores before and after perturbing the test point through (i) label flipping and (ii) image transformation. 
As ablation studies in~\Cref{sec:ablation} show, label flipping is essential for breaking the association between poisoned data and the affected test point, while image transformations introduces randomness that reduces false positive rates by preserving the influence of benign data.

To assess the effectiveness of \delinfl~and the broader applicability of influence functions in this context, we apply our method to three prominent data poisoning attacks: Frequency Trigger~\citep{zeng_rethinking_2021}, Witches' Brew~\citep{geiping_witches_2021}, and BadNet~\citep{gu_2017_badnets}. 
We compare our approach against multiple defenses~\citep{chen_detecting_2018, tran_spectral_2018, zeng_rethinking_2021, grosse_studying_2023, park_2023_trak} that operate with similar or less information about the poisoning than \delinfl. Each attack presents unique challenges for detection and mitigation, as evidenced by the varying performance of existing detection methods across different attacks. 
Additionally, we conduct experiments using known unlearning algorithms to unlearn the poisoning attack using the identified set~\citep{golatkar_2020_ntk, chundawat_2023_can}. 
These experiments provide a comprehensive comparison of these unlearning algorithms. For example, gradient ascent-based methods like SCRUB~\citep{kurmanji_2023_towards} and weight deletion methods like SSD~\citep{foster_2023_fast} can effectively unlearn poisoning when the detected set of training poisons is reasonably accurate. However, their resultant accuracy drops significantly if the detected set includes many falsely flagged clean examples. In contrast, methods like EU and CF~\citep{goel_corrective_2024} are surprisingly robust to false positives, delivering the best unlearning and accuracy. Overall, our experiments demonstrate that \delinfl~consistently outperforms existing algorithms across all settings, offering a robust defense against sophisticated data poisoning attacks while preserving accuracy.

\section{Using Influence functions to detect poisons}
\label{sec:method}

\begin{figure*}[t]
\vspace{-0.5cm}
\begin{center}
%\framebox[4.0in]{$\;$}
% \fbox{\rule[-.5cm]{0cm}{4cm} \rule[-.5cm]{4cm}{0cm}}
\includegraphics[width=0.9\linewidth]{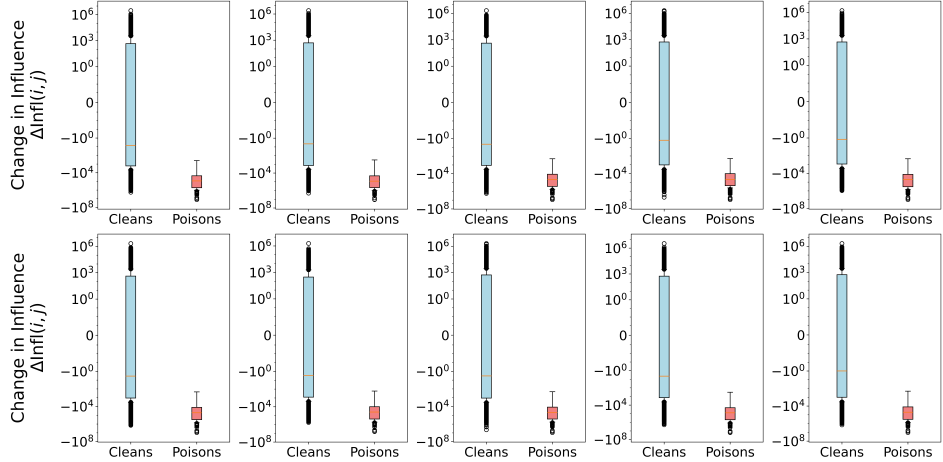}
\end{center}
\vspace{-0.4cm}
\caption{We show the Influence Score Change~(\(\deltainflij{i}{j}\)) for 125 poisoned training points (orange) and 49,875 clean training points (light blue) on the Smooth Trigger attack with CIFAR100. Each plot shows the influence score change for a different transformation applied to the affected test image. Our result shows a consistent drop in influence scores for all poisoned examples after transformation, while clean examples exhibit no clear trend.}
\vspace{-0.4cm}
\label{fig:infl-collapse}
\end{figure*}
In this section, we present how influence functions can be leveraged to unlearn data poisoning attacks and introduce our primary algorithm, \delinfl.\vspace{0.1cm}

Consider a scenario where an adversary modifies a subset of training images belonging to a specific \textit{victim} class by adding a subtle patch or trigger and altering their labels to a \textit{target} class. These manipulated examples, referred to as \textit{poisons}, are incorporated into the training dataset. Consequently, the trained model learns to misclassify any test image from the victim class containing the trigger as belonging to the target class, while maintaining normal performance on other test images.\vspace{0.1cm}

Influence functions~\citep{koh_understanding_2017} provide a mechanism to quantify the contribution of each training example to a particular prediction. By computing the influence of each training point on the prediction of the affected test point, we can identify the most influential training samples responsible for abnormal model behavior. Specifically,  poisoned examples typically exert a significant influence on the affected test predictions, it makes it possible to distinguish the poisons through their influence scores. Thus, influence functions offer a natural approach to trace poisoned training data from misclassified test examples back to the responsible training instances.\vspace{0.1cm}

However, our experiments in~\Cref{sec:exp}, along with several recent
studies~\citep{nguyen_2024_bayesian, bae_2024_training, li2024influence},
demonstrate that naively applying state-of-the-art influence
functions~\citep{grosse_studying_2023} fails to accurately identify
poisoned points in deep neural networks. This limitation necessitates
the development of a more robust method to effectively utilize
influence functions for detecting and unlearning data poisoning.

\subsection{Our Algorithm: $\Delta$-Influence}

To address the shortcomings of the naive approach, we introduce \delinfl.
The core idea is to monitor the changes in influence scores of training data points when the affected test point undergoes various transformations. 
% (See Appendix~\ref{sec:appendix_dont_use_poisoned_train_sample} for a detailed justification regarding the problem setting).

\noindent \textcolor{softred}{\textbf{Notations.}} Let \(\trainz^i \coloneqq \br{\trainx^i, \trainy^i}\) denote a labeled training data point, and let \(\theta^\star\) represent the trained model parameters optimized on the training dataset. For a given test point \(\testz \coloneqq \br{\testx, \testpredy}\) with predicted label \(\testpredy\), the influence function quantifying the impact of \(\trainz^i\) on the loss of \(\testz\) is:

\begin{equation}
    \infl\br{\theta^\star, \trainz^i, \testz} = -\nabla_{\theta} \mathcal{L}\br{\testz, \theta^{\star}}^\top \mathbf{H}^{-1} \nabla_{\theta} \mathcal{L}\br{\trainz^i, \theta^{\star}},
\end{equation}

\noindent where \(\mathcal{L}(z, \theta^\star)\) is the loss evaluated at the
point \(z\) with parameters \(\theta^\star\) and \(\mathbf{H}\) is the
Hessian of the loss function with respect to \(\theta\) at
\(\theta^\star\).\vspace{0.1cm}

\noindent \textcolor{softred}{\textbf{Monitoring Change in Influence.}} Our goal is to attribute the predicted label \(\testpredy\) of a
poisoned test point \(\testz\) to a subset of training points
\(\mathcal{P} = \{\trainz^{1}, \dots, \trainz^{k}\}\) responsible for
the misclassification. To achieve this, \delinfl~monitors the change
in influence scores \(\infl\br{\theta, \trainz^i, \testz}\) for each
training data point \(\trainz^i\) when the test point \(\testz\)
undergoes a set of transformations.\vspace{0.1cm}

Formally, let \(g_j\) be a transformation applied to the test point \(\testz = \br{\testx, \testy}\), consisting of pairing the test image with a random label \(\testy^{\prime}\) 
and applying standard data augmentations such as blurring, color jitter and rotating to \(\testx\) (see Appendix~\ref{sec:predefined_transformations} for the list of all transformations). 
Note that we utilize common data augmentation techniques without designing any poison-specific transformations, suggesting the broad applicability of $\Delta$-Influence. 
We consider such simplicity to be a key strength of our contribution. Then, for each transformation \(g_j\), we compute the change in influence score as:

\begin{equation}\label{eq:infl-delta-eq}
    \begin{aligned}
    \Delta\infl(\theta, {\trainz}^i, g_j\br{\testz}) = &\infl\br{\theta, {\trainz}^i, g_j(\testz) }\\&- \infl\br{\theta, {\trainz}^i, \testz}.
    \end{aligned}
\end{equation}

For brevity, we denote this change as \(\deltainflij{i}{j}\), where \(i\) and \(j\) index the training point and the transformation function, respectively.\vspace{0.1cm}

\noindent \textcolor{softred}{\textbf{Influence Collapse.}} Computing the $\Delta$- Influence is motivated by the following two observations, which we refer to as \emph{Influence Collapse}. Let \(\testz\) be the affected test point.\vspace{0.1cm}

\begin{enumerate}[noitemsep, nolistsep]
    \item \textbf{Negative Change for Poisons:} For all manipulated training samples \(\trainz^i \in \mathcal{P}\) and transformations \(g_j\), the change in influence \(\deltainflij{i}{j}\) is consistently negative.
    \item \textbf{Minimal Change for Clean:} For all clean training samples \(\trainz^k \notin \mathcal{P}\) and transformations \(g_j\), the change in influence \(\deltainflij{k}{j}\) is significantly smaller in magnitude and often positive in value, for most transformations.
\end{enumerate}\vspace{0.1cm}

\noindent This is illustrated in~\Cref{fig:infl-collapse}, where \(\deltainflij{i}{j}\) is consistently negative for poisoned samples across all transformations, whereas it often remains near zero~(compared to that of poisons) or shows no clear trend for clean examples. However,~\Cref{fig:infl-collapse} shows that this is not consistently the case for all clean examples~(with some values being considerably small), which brings us to the next component.\vspace{0.1cm}

\noindent \textcolor{softred}{\textbf{Boosting Using Multiple Transformations.}} 
The above discussion shows that the change in influence score \(\deltainflij{i}{j}\) can be used as a score function for detecting whether \(\trainz^i\) is a manipulated training sample. However, this score function is a relatively weak classifier, especially for clean points, as seen in~\Cref{fig:infl-collapse}. To overcome this problem, we use classical ideas from bagging and apply multiple transformations \(g_1, \ldots, g_{n_b}\) to obtain a series of weak score functions. Specifically, we use \(n_b\) transformations to obtain \(n_b\) weak score functions.\vspace{0.1cm}

Upon obtaining the scores \(\deltainflij{i}{j}\) for all
transformations \(j=1\) to \(j=n_b\), we combine the scores using a
count-based decision rule. Specifically, if a sufficiently large number (\(n - n_{\mathrm{tol}}\)) of transformations lead to a negative change in influence score larger than a threshold \(\tau\), we flag the training data as manipulated. The key hypothesis we leverage here is that for most clean points (unlike the manipulated data), a few transformations will always result in a change in influence score lesser than the threshold \(\tau\). Here \(\tau\) and \(n_{\mathrm{tol}}\) are two hyper-parameters, see Appendix~\ref{sec:appendix_detection_hparams} for details on how to choose them.\vspace{0.1cm}

\noindent \textcolor{softred}{\textbf{Unlearning identified points.}} Once the
set of poisoned training points \(\mathcal{P}\) is identified using
\delinfl, the next step is to unlearn these points to mitigate the
data poisoning attack. We employ several unlearning
algorithms~\citep{goel_2023_adversarial, chundawat_2023_can,
kurmanji_2023_towards, foster_2023_fast} to remove the influence of \(\mathcal{P}\) from the
trained model \(\theta^\star\). In practice, the choice of unlearning
algorithm may depend on factors such as computational efficiency,
scalability, and the specific characteristics of the poisoning attack.
In this work, we look at several popular algorithms including
retraining from scratch~(denoted as EU~\citep{goel_2023_adversarial}),
CF~\citep{goel_2023_adversarial}, SSD~\citep{foster_2023_fast},
SCRUB~\citep{kurmanji_2023_towards}, and BadT~\citep{chundawat_2023_can}.

\subsection{Full Algorithm}

To summarise, the full pipeline of detection and unlearning in
\delinfl~proceeds as follows:

\begin{enumerate}[noitemsep, nolistsep,leftmargin=0.1in]
    \item \textcolor{kleinblue}{Initialization} Begin with trained
    model \(\theta^\star\), a poisoned test point \(\testz\), and the
    entire training dataset \(\mathcal{D} = \{\trainz^i\}_{i=1}^{N}\).
    
    \item \textcolor{kleinblue}{Transformations} Apply a diverse set
    of transformations \(\mathcal{G} = \{g_j\}_{j=1}^{n_b}\) to the
    poisoned test point \(\testz\) to obtain multiple \(\testz' = g_j(\testz)\).
    
    \item \textcolor{kleinblue}{Influence Score} For each training data point \(\trainz^i \in \mathcal{D}\) and each transformation \(g_j \in \mathcal{G}\), compute the change in influence score \(\deltainflij{i}{j}\) as defined in~\Cref{eq:infl-delta-eq}.
    
    \item \textcolor{kleinblue}{Boosting and Detection} For each training data point \(\trainz^i\), aggregate the influence score changes across all transformations. If the number of significant negative changes (\(\deltainflij{i}{j} < \tau\)) exceeds \(n_b - n_{\mathrm{tol}}\), flag \(\trainz^i\) as a poisoned sample.
    %If the number of negative changes exceeds a predefined threshold \(\tau\), flag \(\trainz^i\) as a poisoned sample.
    
    \item \textcolor{kleinblue}{Unlearning} Once the set of poisoned training points \(\mathcal{P}\) is identified, apply unlearning algorithms to remove their influence from the trained model \(\theta^\star\).
\end{enumerate}

In the next section, we use the above algorithm for experiments on
several data poisons, datasets, and unlearning algorithms and compare
them with existing approaches.

\section{Experiments}\label{sec:exp}

We now showcase the empirical performance of our algorithm in comparison to multiple baselines.
\subsection{Experimental Setup}
\textcolor{softred}{\textbf{Attacks.}} To ensure broad coverage and robustness, we evaluate our \delinfl~algorithm against three distinct types of data poisoning attacks:

\begin{enumerate}[leftmargin=*]
    \item \textcolor{kleinblue}{{Frequency Trigger
    ~\citep{zeng_rethinking_2021}}}: In this approach, along with changing the label, a trained,
    imperceptible pattern is embedded in both the spatial
    and frequency domains, thereby encompassing the whole image. 
    As shown in~\citet{alex_2024_protecting}, these patterns are difficult to detect by both human and
    automated methods, making the poisoned samples
    challenging to identify.
    
    \item \textcolor{kleinblue}{{Clean Label Attack
    (Witches' Brew) ~\citep{geiping_witches_2021}}}: Unlike Frequency Trigger,
    this attack adds an imperceptible pattern to images without
    altering their labels. The poisoned samples appear benign since
    their labels are consistent with their content, yet they cause the
    model to learn incorrect associations, leading to
    misclassifications during inference. As shown
    in~\citet{pawelczyk_machine_2024}, these patterns are difficult to
    unlearn using unlearning algorithms.

    \item \textcolor{kleinblue}{{Patch Trigger (BadNet)
    ~\citep{gu_2017_badnets}}}: Also studied
    in~\citet{goel_corrective_2024}, this attack involves adding a subtle patch to the corner of selected training images and
    altering their labels to a designated target class. The presence
    of the patch causes the model to misclassify any test
    image containing the patch into the target class while maintaining
    normal performance on other inputs.
\end{enumerate}

% \begin{figure*}[t]
%     \centering
%     \vspace{-0.5cm}
%     \includegraphics[width=0.7\linewidth]{files/resulta.pdf}
%     \vspace{-0.15cm}
%     \caption{\textcolor{kleinblue2}{\underline{Size of Discarded Clean Samples per Method:}} We evaluate how many clean samples are mistakenly discarded during poison detection, impacting  accuracy (model utility). SpecSig, ActClust, and EK-FAC perform poorly (larger circle sizes indicate higher discard rates), while FreqDef and $\Delta$-Influence are better at preserving clean data, though both have low true positive rates and room for improvement in detection efficacy.}
%     \vspace{-0.4cm}
%     \label{fig:tpsize}
    
% \end{figure*}

\noindent \textcolor{softred}{\textbf{Model and Datasets.}} We utilize the CIFAR10 and CIFAR100 datasets~\citep{Krizhevsky_2009_LearningML} and a ResNet18 model~\citep{he_deep_2015}, following the standard benchmarks and models used in the state-of-the-art machine unlearning setup~\citep{pawelczyk_machine_2024}. For CIFAR10, we poison 500 training images (1\% of the dataset), while for CIFAR100, we poison 125 training images (0.25\% of the dataset) for all attack types except BadNet, which requires a higher size of 350 samples to be effective. The victim class and attack class (when different) are selected randomly. Detection methods are tuned on a small validation set using cross-validation techniques. Hyperparameters such as threshold values and clustering parameters are optimized based on validation performance metrics to achieve the best balance between detection accuracy and false positive rates. 
Detailed hyperparameter settings and our code are provided in the Appendix~\ref{sec:appendix_exp_details} to ensure reproducibility.

\noindent \textcolor{softred}{\textbf{Compared Methods.}} We compare the
detection performance of existing popular methods in the data
poisoning literature by adapting them to our setting. Additionally, we
include the state-of-the-art methods for computing influence function:
EK-FAC~\citep{grosse_studying_2023} and TRAK~\citep{park_2023_trak} as baselines. Our \delinfl~method
is built upon EK-FAC. 

\begin{enumerate}[leftmargin=*]
    \item \textcolor{kleinblue}{{Activation
    Clustering-Based Detection \citep{chen_detecting_2018}}}
    identifies backdoored samples by clustering the activations of the
    last hidden layer for each class. If a class's activations can be
    effectively clustered into two distinct groups, the smaller
    cluster is deemed to contain poisoned samples and is subsequently
    removed for retraining.
    
    \item \textcolor{kleinblue}{{Spectral Signature-Based
    Detection \citep{tran_spectral_2018}}} employs singular value
    decomposition on the activations of the last hidden layer per
    class. Samples with high values in the first singular dimension
    are flagged as poisoned and removed based on a predefined
    hyperparameter threshold.
    
    \item \textcolor{kleinblue}{{Frequency-Based Detection
    \citep{zeng_rethinking_2021}}} performs frequency analysis by
    building a classifier on the discrete cosine transforms of
    synthetic images containing hardcoded backdoor-like features. It
    identifies poisoned examples by detecting these frequency-based
    patterns.
    
    \item \textcolor{kleinblue}{{EK-FAC
    \citep{grosse_studying_2023}}}  serves as our baseline method for
    using influence functions in poison detection. It calculates
    influence scores for every training sample based on one known
    affected test sample. Samples with average scores exceeding
    a predefined threshold are removed.

    \item \textcolor{kleinblue}{{TRAK
    \citep{park_2023_trak}}} uses another implementation of influence functions when thresholding.
\end{enumerate}

\noindent\textcolor{softred}{\textbf{Metrics.}} We evaluate our
algorithm using five key metrics. All metrics are averaged over three random seeds.
\begin{enumerate}[leftmargin=*]
    \item \textcolor{kleinblue}{{True Positive Rate (TPR):}}
    Fraction of identified poisoned samples out of the total
    poisoned samples in train set.
    \[\footnotesize \frac{\text{Number of correctly flagged poisoned samples}}{\text{Total number of poisoned samples}} \times 100\%
    \]
    
    \item \textcolor{kleinblue}{{Precision:}} Proportion of correctly identified poisoned samples among all flagged samples. It captures the trade-off between detection accuracy and model utility.
    \[\footnotesize \frac{\text{Number of correctly flagged poisoned samples}}{\text{Total number of samples flagged as poisoned}} \times 100\%
    \]

    \item \textcolor{kleinblue}{{Poison Success Rate (PSR):}}
    Fraction of poisoned test samples that are misclassified into the
    target (incorrect) class.
    \[\footnotesize \frac{\text{Number of poisoned samples classified as target}}{\text{Total number of poisoned samples}} \times 100\%
    \]
    
    \item \textcolor{kleinblue}{{Test Accuracy:}} The performance on unpoisoned test samples, measuring drop in model utility.
    \[\footnotesize \frac{\text{Number of correct predictions on test set}}{\text{Total number of test samples}} \times 100\%
    \]

    \item \textcolor{kleinblue}{{Area under the ROC curve (AuROC):}} Trade-off between TPR and Precision due to the choice of threshold.

\end{enumerate}

\begin{figure}[t]
    \centering
    % \vspace{-1cm}
    \includegraphics[width=\linewidth]{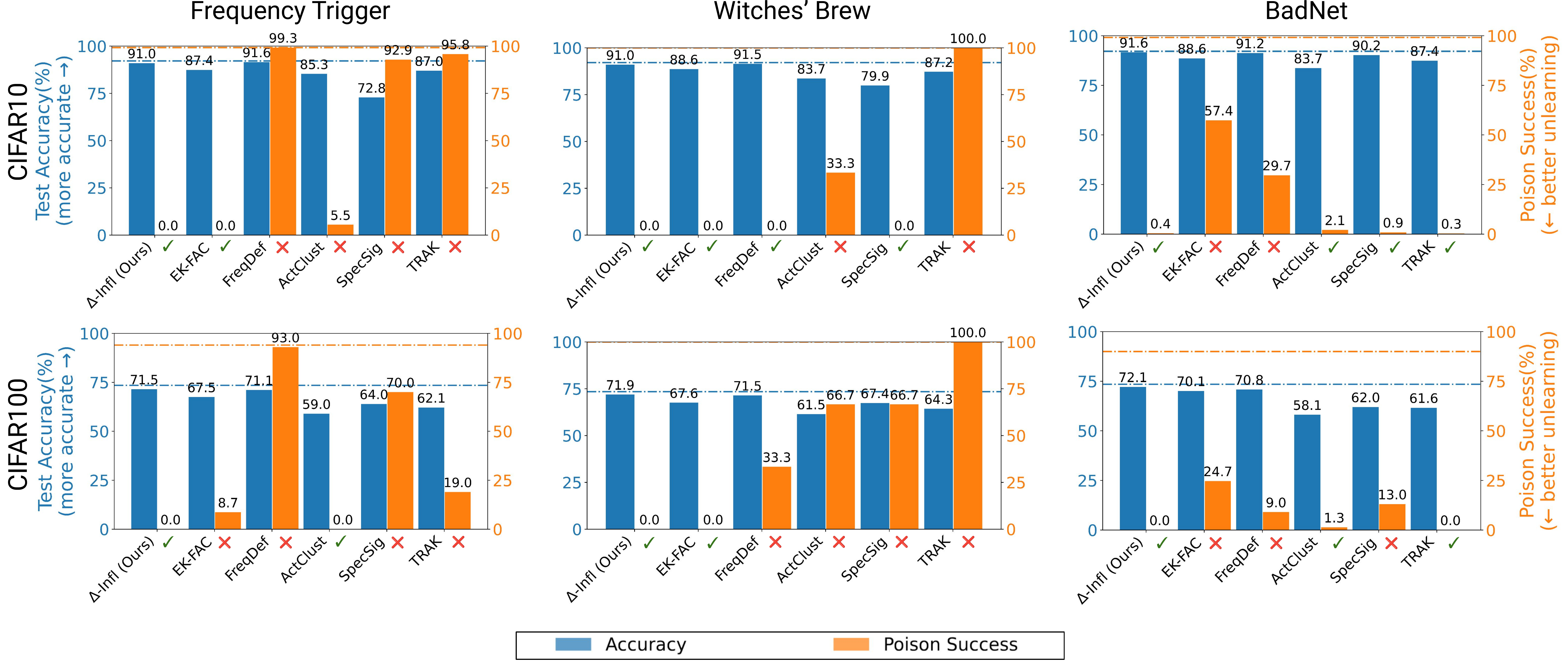}
    \caption{\textcolor{kleinblue2}{\underline{Poison Success Rate}} and \textcolor{kleinblue2}{\underline{Test Accuracy.}} This table shows both poison unlearning effectiveness and model utility. A method is considered successful if the poison success rate is below 5\%, marked by \textcolor{softgreen}{\checkmark}, with unsuccessful methods marked by \textcolor{softred}{\textbf{$\times$}}. $\Delta$-Influence is successful in 6/6 cases, while the closest competitors succeed in only 3/6. Additionally, $\Delta$-Influence nearly perfectly preserves test accuracy. Figure structure from \citep{pawelczyk_machine_2024}.}
    \vspace{-0.25cm}
    \label{fig:unlearningperf}
\end{figure}

\begin{table}[!t]
    \centering
    % \small
    % \footnotesize
    \scriptsize
    \resizebox{0.91\linewidth}{!}{
    \begin{tabular}{l l c c c c c c}
    \toprule
    \textbf{Method} & \textbf{Metric} & \multicolumn{3}{c}{\textbf{CIFAR10}} & \multicolumn{3}{c}{\textbf{CIFAR100} } \\
     & & {Frequency} & {Witches'} & {BadNet} & {Frequency} & {Witches'} & {BadNet} \\
     & & Trigger & Brew & & Trigger & Brew & \\
    \midrule
    \multirow{2}{*}{SpecSig} & Precision & \textcolor{softred}{1.3\%} & \textcolor{softgreen}{1.4\%} & \textcolor{softgreen}{3.6\%} & \textcolor{softred}{0.5\%} & \textcolor{softred}{0.3\%} & \textcolor{softred}{1.3\%} \\
                                       & TPR       & \textcolor{softred}{88.3\%} & \textcolor{softgreen}{96.8\%} & \textcolor{softgreen}{88.3\%} & \textcolor{softred}{78.4\%} & \textcolor{softred}{35.2\%} & \textcolor{softred}{82.6\%} \\
                                       & AuROC       & \textcolor{softred}{0.53} & \textcolor{softgreen}{0.88} & \textcolor{softgreen}{0.82} & \textcolor{softred}{0.72} & \textcolor{softred}{0.58} & \textcolor{softred}{0.76} \\
    \addlinespace[0.5em] % Add space between methods
    \multirow{2}{*}{ActClust}   & Precision & \textcolor{softred}{2.2\%} & \textcolor{softred}{2.1\%} & \textcolor{softgreen}{2.2\%} & \textcolor{softgreen}{0.6\%} & \textcolor{softred}{0.3\%} & \textcolor{softgreen}{1.6\%} \\
                                       & TPR       & \textcolor{softred}{99.1\%} & \textcolor{softred}{93.4\%} & \textcolor{softgreen}{94.9\%} & \textcolor{softgreen}{100\%} & \textcolor{softred}{55.2\%} & \textcolor{softgreen}{96.3\%}\\
                                       & AuROC       & \textcolor{softred}{0.77} & \textcolor{softred}{0.75} & \textcolor{softgreen}{0.79} & \textcolor{softgreen}{0.79} & \textcolor{softred}{0.53} & \textcolor{softgreen}{0.78} \\
    \addlinespace[0.5em] % Add space between methods
    \multirow{2}{*}{FreqDef}   & Precision & \textcolor{softred}{0.4\%} & \textcolor{softgreen}{10.2\%} & \textcolor{softred}{8.0\%} & \textcolor{softred}{0.1\%} & \textcolor{softred}{1.8\%} & \textcolor{softred}{5.3\%} \\
                                       & TPR       & \textcolor{softred}{3.2\%} & \textcolor{softgreen}{93.6\%} & \textcolor{softred}{72.3\%} & \textcolor{softred}{2.4\%} & \textcolor{softred}{78.4\%} & \textcolor{softred}{85.7\%} \\
                                       & AuROC       & \textcolor{softred}{0.32} & \textcolor{softgreen}{0.98} & \textcolor{softred}{0.97} & \textcolor{softred}{0.28} & \textcolor{softred}{0.91} & \textcolor{softred}{0.97} \\
    \addlinespace[0.5em] % Add space between methods
    \multirow{2}{*}{TRAK}   & Precision & \textcolor{softred}{1.4\%} & \textcolor{softred}{1.0\%} & \textcolor{softgreen}{1.9\%} & \textcolor{softred}{0.5\%} & \textcolor{softred}{0.2\%} & \textcolor{softgreen}{1.4\%} \\
                                       & TPR       & \textcolor{softred}{70.6\%} & \textcolor{softred}{49.8\%} & \textcolor{softgreen}{93.6\%} & \textcolor{softred}{96.8\%} & \textcolor{softred}{48.0\%} & \textcolor{softgreen}{100\%} \\
                                       & AuROC       & \textcolor{softred}{0.73} & \textcolor{softred}{0.50} & \textcolor{softgreen}{0.60} & \textcolor{softred}{0.79} & \textcolor{softred}{0.49} & \textcolor{softgreen}{0.58} \\
    \addlinespace[0.5em] % Add space between methods
    \multirow{2}{*}{EK-FAC}   & Precision & \textcolor{softgreen}{2.9\%} & \textcolor{softgreen}{0.8\%} & \textcolor{softred}{2.8\%} & \textcolor{softred}{0.9\%} & \textcolor{softgreen}{0.4\%} & \textcolor{softred}{3.2\%} \\
                                       & TPR       & \textcolor{softgreen}{100\%} & \textcolor{softgreen}{17.4\%} & \textcolor{softred}{67.1\%} & \textcolor{softred}{96.8\%} & \textcolor{softgreen}{47.2\%} & \textcolor{softred}{70.0\%} \\
                                       & AuROC       & \textcolor{softgreen}{0.89} & \textcolor{softgreen}{0.57} & \textcolor{softred}{0.87} & \textcolor{softred}{0.94} & \textcolor{softgreen}{0.68} & \textcolor{softred}{0.71} \\
    \addlinespace[0.25em] % Add space between methods
    \cellcolor{gray!10}                     & \cellcolor{gray!10}\rule{0pt}{1.0em}Precision & \cellcolor{gray!10}\textcolor{softgreen}{13.3\%} & \cellcolor{gray!10}\textcolor{softgreen}{3.3\%} & \cellcolor{gray!10}\textcolor{softgreen}{17.6\%} & \cellcolor{gray!10}\textcolor{softgreen}{2.9\%} & \cellcolor{gray!10}\textcolor{softgreen}{2.1\%} & \cellcolor{gray!10}\textcolor{softgreen}{37.3\%} \\
                                        \cellcolor{gray!10}$\Delta$-Infl (Ours) & \cellcolor{gray!10}TPR & \cellcolor{gray!10}\textcolor{softgreen}{100\%} & \cellcolor{gray!10}\textcolor{softgreen}{19.4\%} & \cellcolor{gray!10}\textcolor{softgreen}{99.1\%} & \cellcolor{gray!10}\textcolor{softgreen}{100\%} & \cellcolor{gray!10}\textcolor{softgreen}{62.4\%} & \cellcolor{gray!10}\textcolor{softgreen}{96.9\%} \\
                                        \cellcolor{gray!10}                     & \cellcolor{gray!10}AuROC & \cellcolor{gray!10}\textcolor{softgreen}{0.96} & \cellcolor{gray!10}\textcolor{softgreen}{0.38} & \cellcolor{gray!10}\textcolor{softgreen}{0.95} & \cellcolor{gray!10}\textcolor{softgreen}{0.96} & \cellcolor{gray!10}\textcolor{softgreen}{0.75} & \cellcolor{gray!10}\textcolor{softgreen}{0.82} \\
    \bottomrule
    \end{tabular}}
    \vspace{0.5em}
    \caption{Comparison of \textcolor{kleinblue2}{\underline{Precision}}~\&~\textcolor{kleinblue2}{\underline{TPR}}~\&~\textcolor{kleinblue2}{\underline{AuROC}} across methods and dataset for detecting poisoned samples. \textcolor{softgreen}{Green} indicates successful unlearning (PSR \(\leq 5\%\), while \textcolor{softred}{red} indicates failed unlearning (see \Cref{fig:unlearningperf} for exact poisoning success rates). We evaluate the precision and TPR of detecting poisoned training samples. SpecSig \cite{tran_spectral_2018}, ActClust \cite{chen_detecting_2018}, TRAK \cite{park_2023_trak} and EK-FAC \cite{grosse_studying_2023} yield low precision, flagging many clean samples as poisoned. FreqDef \cite{zeng_rethinking_2021} and \delinfl better preserve clean data, though FreqDef shows a significantly lower TPR, missing many true poisoned samples. For BadNet, the poisoning success rate correlates with the number of detected poisoned samples, making the attack in \cite{goel_corrective_2024} relatively easy to unlearn. In contrast, the Frequency attack requires nearly all poisoned samples to be removed for recovery, making it particularly challenging. Surprisingly, the Witches' Brew setting is easier than anticipated \cite{pawelczyk_machine_2024}, requiring only a few key samples—mainly identified by influence functions—to be removed for effective unlearning.}
    \label{tab:cifar_precision_tpr}
\end{table}

\subsection{Main Results}

\hspace{0.4cm}We present our experimental findings across the above metrics and
compare the performance of \delinfl~against several baselines.
Specifically, we report the  precision, TPR, and AuROC of detecting poisons in
\Cref{tab:cifar_precision_tpr}, and the overall PSR and test accuracy after retraining without the identified set in \Cref{fig:unlearningperf}.\vspace{0.1cm}

\noindent \textcolor{softred}{\textbf{Performance of
$\Delta$-Influence.}} As illustrated in \Cref{fig:unlearningperf}, with~\textcolor{kleinblue2}{{EU~\citep{goel_2023_adversarial}}} as the unlearning method, 
\delinfl consistently achieves a poison success rate below
5\% across all three types of poisoning attacks and both datasets.
This success rate is marked by a \textcolor{softgreen}{\checkmark},
while unsuccessful detections are marked by a
\textcolor{softred}{\textbf{$\times$}}. In contrast, the next best
methods, Activation Clustering (ActClust) and EK-FAC, succeed in only
3 out of 6 cases, as highlighted in \Cref{tab:cifar_precision_tpr}. This
showcases the substantial improvement in performance gained by
\delinfl. We further validate the effectiveness and robustness of \delinfl across different unlearning methods, \textit{e.g.}, a popular alternative unlearning algorithm called~\textcolor{kleinblue2}{{SCRUB
~\citep{kurmanji_2023_towards}}} which involves gradient ascent. See detailed results in~\Cref{fig:scrub_unlearn} in Appendix.

Among the baseline methods, EK-FAC outperforms ActClust by minimizing
the drop in test accuracy, also indicated by a higher precision in \Cref{tab:cifar_precision_tpr}. Furthermore,
$\Delta$-Influence consistently achieves the highest precision, offering better performance with minimal accuracy loss compared to the
other methods. Additional experiments detailed in~\Cref{sec:perturb_only_images_or_labels}
demonstrate that both label and input augmentations are necessary for~\delinfl.

\noindent \textcolor{softred}{\textbf{Variance across Poisons.}} Our
analysis shows that the BadNet poison can be effectively removed
without identifying all poisoned samples, reaffirming  that it is
realatively easy to eliminate. Based on these results, we advocate
that the corrective unlearning literature should benchmark proposed
algorithms on the more challenging frequency-based
poisons~\citep{zeng_rethinking_2021}, which require detecting nearly
all poisoned samples and are notably harder to remove with a partial
subset. This was also identified to be difficult in previous work \citep{alex_2024_protecting}.

Surprisingly, in the case of the Witches' Brew attack on CIFAR-10,
our~\delinfl method often identifies fewer but a sufficient number of
true poisoned samples compared to other methods. We attribute this to
the unique behavior of this particular poison.~\delinfl effectively
identifies the samples most responsible for the misclassification, and
in Witches' Brew, only a few samples are truly effective for
poisoning. Additional experiments in~\Cref{sec:counterfactual_analysis} show that removing the
complement of detected poisons does not allow the model to recover,
despite the complement set being similar in size or larger.

\noindent \textcolor{softred}{\textbf{Conclusion.}} Overall,~\delinfl offers an effective mechanism for unlearning data
poisonining attacks without significantly impacting model performance. 
Crucially, it requires \emph{no prior knowledge of the attack method}, making it more generalizable across various poisoning strategies.

\section{Unpacking Key Factors in $\Delta$-Infleunce}
\label{sec:ablation}
We present a series of additional analyses designed
to improve the understanding of \delinfl.
Specifically, we explore: (i) individual contributions of image and
label perturbations, (ii) effectiveness of various unlearning
algorithms, (iii) a counterfactual analysis to determine whether
the detected samples are solely responsible for enabling poisoning in
the Witches' Brew attack, and (iv) the unreliability of using a known training poison as an attribution target.

\subsection{Perturbing Only Images or Labels}
\label{sec:perturb_only_images_or_labels}

\begin{table}[t]
    \centering
    \footnotesize
    \resizebox{0.8\linewidth}{!}{
    \begin{tabular}{l l c c c c c c}
    \toprule
    \textbf{Method} & \textbf{Metric} & \multicolumn{3}{c}{\textbf{CIFAR10}} & \multicolumn{3}{c}{\textbf{CIFAR100} } \\
     & & {Frequency} & {Witches'} & {BadNet} & {Frequency} & {Witches'} & {BadNet} \\
     & & Trigger & Brew & & Trigger & Brew & \\
    \midrule
    \multirow{2}{*}{Ours (Label-Only)} & Precision & \textcolor{softgreen}{6.3\%} & \textcolor{softgreen}{1.2\%} & \textcolor{softgreen}{4.0\%} & \textcolor{softgreen}{1.1\%} & \textcolor{softgreen}{0.8\%} & \textcolor{softgreen}{3.1\%} \\
                                       & TPR       & \textcolor{softgreen}{100\%} & \textcolor{softgreen}{24.2\%} & \textcolor{softgreen}{97.5\%} & \textcolor{softgreen}{100\%} & \textcolor{softgreen}{73.6\%} & \textcolor{softgreen}{99.1\%} \\
                                       & AuROC       & \textcolor{softgreen}{0.94} & \textcolor{softgreen}{0.48} & \textcolor{softgreen}{0.90} & \textcolor{softgreen}{0.92} & \textcolor{softgreen}{0.69} & \textcolor{softgreen}{0.79} \\
    \addlinespace[0.5em] % Add space between methods
    \multirow{2}{*}{Ours (Img-Only)}   & Precision & \textcolor{softred}{28.9\%} & \textcolor{softred}{2.7\%} & \textcolor{softred}{14.4\%} & \textcolor{softred}{0.6\%} & \textcolor{softred}{0.3\%} & \textcolor{softred}{7.6\%} \\
                                       & TPR       & \textcolor{softred}{26.4\%} & \textcolor{softred}{13.2\%} & \textcolor{softred}{68.9\%} & \textcolor{softred}{62.4\%} & \textcolor{softred}{40.8\%} & \textcolor{softred}{50.6\%} \\
                                       & AuROC       & \textcolor{softred}{0.51} & \textcolor{softred}{0.25} & \textcolor{softred}{0.85} & \textcolor{softred}{0.78} & \textcolor{softred}{0.62} & \textcolor{softred}{0.76} \\
    \addlinespace[0.25em] % Add space between methods
    \cellcolor{gray!10}                     & \cellcolor{gray!10}\rule{0pt}{1.0em}Precision & \cellcolor{gray!10}\textcolor{softgreen}{13.3\%} & \cellcolor{gray!10}\textcolor{softgreen}{3.3\%} & \cellcolor{gray!10}\textcolor{softgreen}{17.6\%} & \cellcolor{gray!10}\textcolor{softgreen}{2.9\%} & \cellcolor{gray!10}\textcolor{softgreen}{2.1\%} & \cellcolor{gray!10}\textcolor{softgreen}{37.3\%} \\
    \cellcolor{gray!10}Ours (Both)       & \cellcolor{gray!10}TPR       & \cellcolor{gray!10}\textcolor{softgreen}{100\%} & \cellcolor{gray!10}\textcolor{softgreen}{19.4\%} & \cellcolor{gray!10}\textcolor{softgreen}{99.1\%} & \cellcolor{gray!10}\textcolor{softgreen}{100\%} & \cellcolor{gray!10}\textcolor{softgreen}{62.4\%} & \cellcolor{gray!10}\textcolor{softgreen}{96.9\%} \\
    \cellcolor{gray!10}                     & \cellcolor{gray!10}AuROC       & \cellcolor{gray!10}\textcolor{softgreen}{0.96} & \cellcolor{gray!10}\textcolor{softgreen}{0.38} & \cellcolor{gray!10}\textcolor{softgreen}{0.95} & \cellcolor{gray!10}\textcolor{softgreen}{0.96} & \cellcolor{gray!10}\textcolor{softgreen}{0.75} & \cellcolor{gray!10}\textcolor{softgreen}{0.82} \\
    \bottomrule
    \end{tabular}}
    \vspace{0.5em}
    \caption{Comparison of \textcolor{kleinblue2}{\underline{Precision}}~\&~\textcolor{kleinblue2}{\underline{TPR}}~\&~\textcolor{kleinblue2}{\underline{AuROC}} across Label-Only, Image-Only and combined transformation of affected image. \textcolor{softgreen}{Green} indicates successful unlearning (PSR \(< 5\%\)), while \textcolor{softred}{red} indicates unsuccessful unlearning (See Appendix for exact PSR). Label-only augmentations are highly effective in detecting poisoned samples, whereas image-only augmentations perform poorly. Conversely, image-only augmentations significantly reduce the FPR, preserving more clean data and improving detection precision.}
    \label{tab:combined_3and4}
\end{table}

\noindent\textcolor{softred}{\textbf{Setup.}} To distinguish the contributions of image and label
perturbations in our \delinfl~method, we conduct an ablation study by
evaluating the two key components separately:

\textcolor{kleinblue}{{1. Modify Label (\delinfl (Label-Only))}}: Conversely, in this baseline, we only modify the test point's labels while keeping the images unchanged. This setup helps evaluate the effect of label manipulation on detecting the influence of poisoned training points.

\textcolor{kleinblue}{{2. Modify Image (\delinfl
(Img-Only))}}: In this baseline, we exclusively modify the test images
without altering their labels. This allows us to isolate the
impact of image transformations on the model's ability to detect
poisoned data.
   
Both ablations are benchmarked across the same datasets and poisoning attacks, utilising identical metrics to ensure consistency in evaluation. The goal is to understand the individual and combined effects of image and label perturbations on the detection performance of \delinfl.

\noindent\textcolor{softred}{\textbf{Results.}} As depicted in \Cref{tab:combined_3and4},
our ablation study reveals that label-only augmentations achieve high
TPR across all poisoning types and datasets, effectively identifying
almost all poisoned samples. However, this leads to low precision, resulting in the
unnecessary removal of a significant number of clean samples. On the
other hand, image-only augmentations exhibit poor TPR, failing at the core task but also rejects less clean
samples (higher precision). In contrast, \delinfl leverages both label and image perturbations to
achieve a balanced performance and detects sufficient key poisoned samples while rejecting lesser clean samples (see \Cref{fig:abl_unlearn} in Appendix for detailed unlearning performance).

\noindent\textcolor{softred}{\textbf{Conclusion.}} Our ablation study underscores the necessity of
incorporating both label and image augmentations in the \delinfl
method. Label flippings are pivotal for enhancing detection
accuracy, while image transformations play a critical role in minimizing
false positives. 

\subsection{Which Unlearning Methods Work?}

\begin{figure*}
    \centering
    \includegraphics[width=\linewidth]{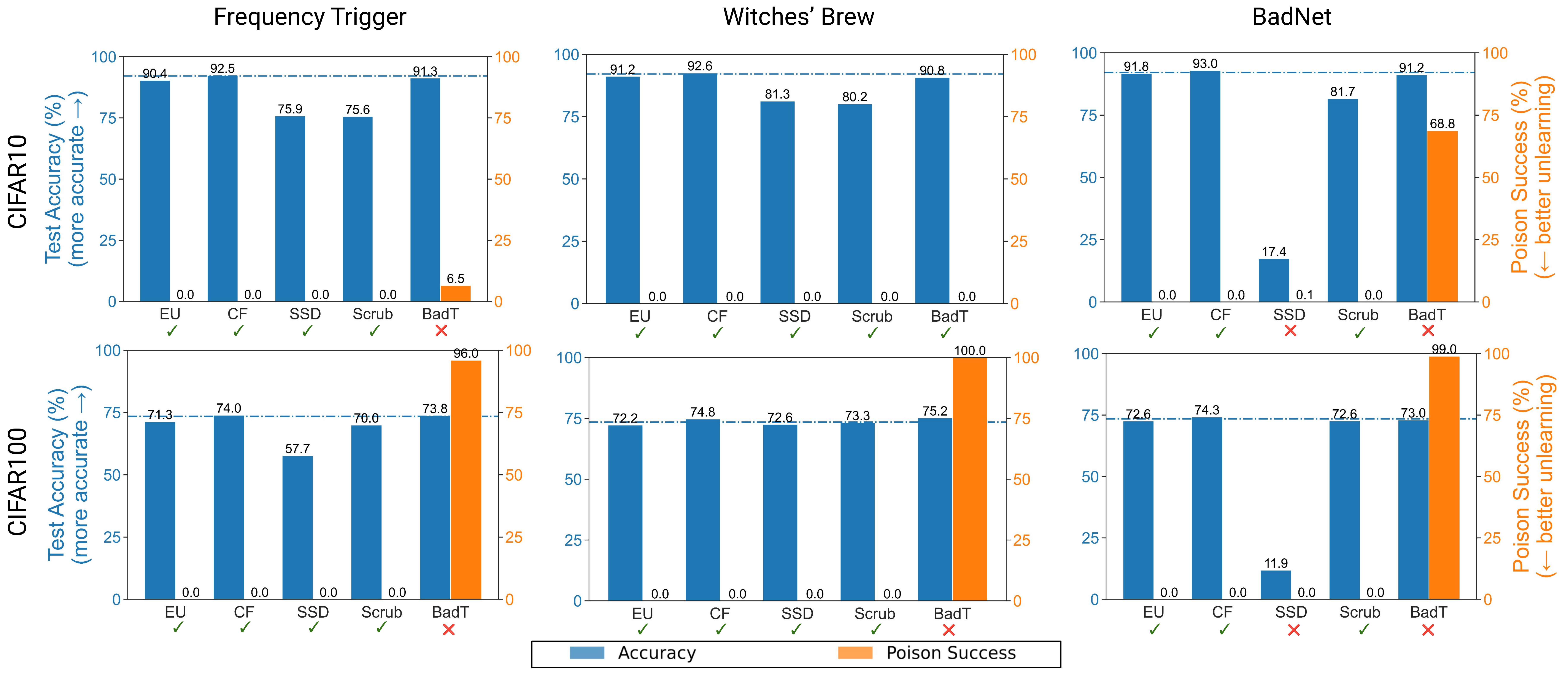}
    \caption{\textcolor{kleinblue2}{Poison Success Rate} and \textcolor{kleinblue2}{Test Accuracy} for Unlearning Methods Applied on Samples Identified by \delinfl. Catastrophic Forgetting (CF) and Exact Unlearning (EU) from \citet{goel_2023_adversarial} perform best, effectively unlearning poisoned samples while maintaining test accuracy. In contrast, SSD~\citep{foster_2023_fast} and SCRUB~\citep{kurmanji_2023_towards} struggle with false negatives, leading to significant accuracy drops, while BadT~\citep{chundawat_2023_can} fails to unlearn effectively. We recommend EU or CF as strong baselines and highlight the need for future methods to improve robustness against false positives.}
    \label{fig:unlearning_method}
\end{figure*}

\noindent\textcolor{softred}{\textbf{Setup.}} To evaluate the effectiveness of various unlearning
algorithms when paired with \delinfl, we fix the influence
function to \delinfl and vary the unlearning functions. We benchmark
several corrective unlearning methods,
including~\textcolor{kleinblue2}{{EU~\citep{goel_2023_adversarial}}},
\textcolor{kleinblue2}{{CF ~\citep{goel_2023_adversarial}}},
\textcolor{kleinblue2}{{SSD
~\citep{foster_2023_fast}}},~\textcolor{kleinblue2}{{SCRUB
~\citep{kurmanji_2023_towards}}}, and~\textcolor{kleinblue2}{{BadT
~\citep{chundawat_2023_can}}}. All methods are implemented using the codebase and training protocols from \citet{goel_corrective_2024}. 
Further implementation details are provided in Appendix~\ref{sec:appendix_exp_details}.

\noindent\textcolor{softred}{\textbf{Results.}} As illustrated in \Cref{fig:unlearning_method}, our
evaluation reveals that CF performs comparably to EU, achieving
similar poison removal success rates while offering significant
computational gains by avoiding full retraining. CF remains robust
against false positives, maintaining high test accuracy. EU
effectively removes poisoned samples with no significant drop in test
accuracy, albeit at a higher computational cost due to retraining. In
contrast, while gradient-ascent-based methods like SCRUB and weight
deletion approaches like SSD successfully unlearn poisons, they do
so at the expense of model utility due to their susceptibility to
false positives. Finally, BadT fails to unlearn poisons effectively.
We recommend EU or CF as
competitive baselines for corrective unlearning using influence functions, and also highlight the importance of robustness towards false positives.

\noindent\textcolor{softred}{\textbf{Conclusion.}} We recommend EU or CF as
competitive baselines for corrective unlearning using influence functions, and also highlight the importance of robustness towards false positives.

\subsection{Counterfactual Analysis: Do Detected Samples Account for Poisoning in Witches' Brew?}
\label{sec:counterfactual_analysis}

\noindent\textcolor{softred}{\textbf{Setup.}} The analysis compares the
original detected set of poisoned samples in Witches' Brew to its complement set (i.e., all poisoned samples except those
detected by \delinfl). This aims to assess whether the
detected set exclusively accounts for the poisoning effect.

\noindent\textcolor{softred}{\textbf{Results.}} As presented in \Cref{tab:counterfact}, the removal
of the ``Original'' detected set (19.4\% TPR for CIFAR10 and 62.4\%
TPR for CIFAR100) results in 0\% poison success rate, effectively
unlearning the poisoning. In stark contrast, removing the
``Complement'' set (80.6\% TPR for CIFAR10 and 37.6\% TPR for
CIFAR100) maintains a poison success rate of 100\%, indicating that
the undetected samples do not sufficiently contribute to the
poisoning. The complement set achieves higher test
accuracy simply because it only contains unaffected samples without
false positives.
These results demonstrate that our detected
subset accounts for nearly all the poisoning effects in Witches' Brew,
highlighting the unusual nature of this particular poison as well as the precision of
our~\delinfl algorithm.

\noindent\textcolor{softred}{\textbf{Conclusion.}} These results demonstrate that our detected
subset accounts for nearly all the poisoning effects in Witches' Brew,
highlighting the unusual nature of this particular poison as well as the precision of
our~\delinfl algorithm.

\begin{table}[h]
\centering
\caption{\textbf{Does the Detected Set Truly Influence the Poison?} For Witches' Brew, we test the ``Original'' set, representing the poisoned samples identified by \delinfl, and the``Complement'' set, which includes all other poisoned samples not detected. The absence of a drop in poison success rate when removing the complement set suggests that the detected set fully captures the poisoning effect. Conversely, removing the detected set completely eliminates the poisoning effect.}
\vspace{1em}
\footnotesize
\label{tab:counterfact}
\resizebox{0.7\linewidth}{!}{
\begin{tabular}{cccc}
\toprule
\textbf{$\Delta$-Influence Set} & \textbf{TPR}($\uparrow$) & \textbf{Poison Success Rate} ($\downarrow$) & \textbf{Test Accuracy} ($\uparrow$) \\
\midrule
\multicolumn{4}{c}{\textbf{CIFAR10}} \\
\midrule
Original        & 19.4\%  & \textcolor{softgreen}{0\%}    & 91.0\%  \\
Complement  & 80.6\% & \textcolor{softred}{100\%}  & 92.2\%  \\
\midrule
\multicolumn{4}{c}{\textbf{CIFAR100}} \\
\midrule
Original         & 62.4\% & \textcolor{softgreen}{0\%}    & 71.9\%  \\
Complement Set  & 37.6\%  & \textcolor{softred}{100\%}  & 72.8\%  \\
\bottomrule
\end{tabular}}
\end{table}

\begin{figure*}
    \includegraphics[width=\linewidth]{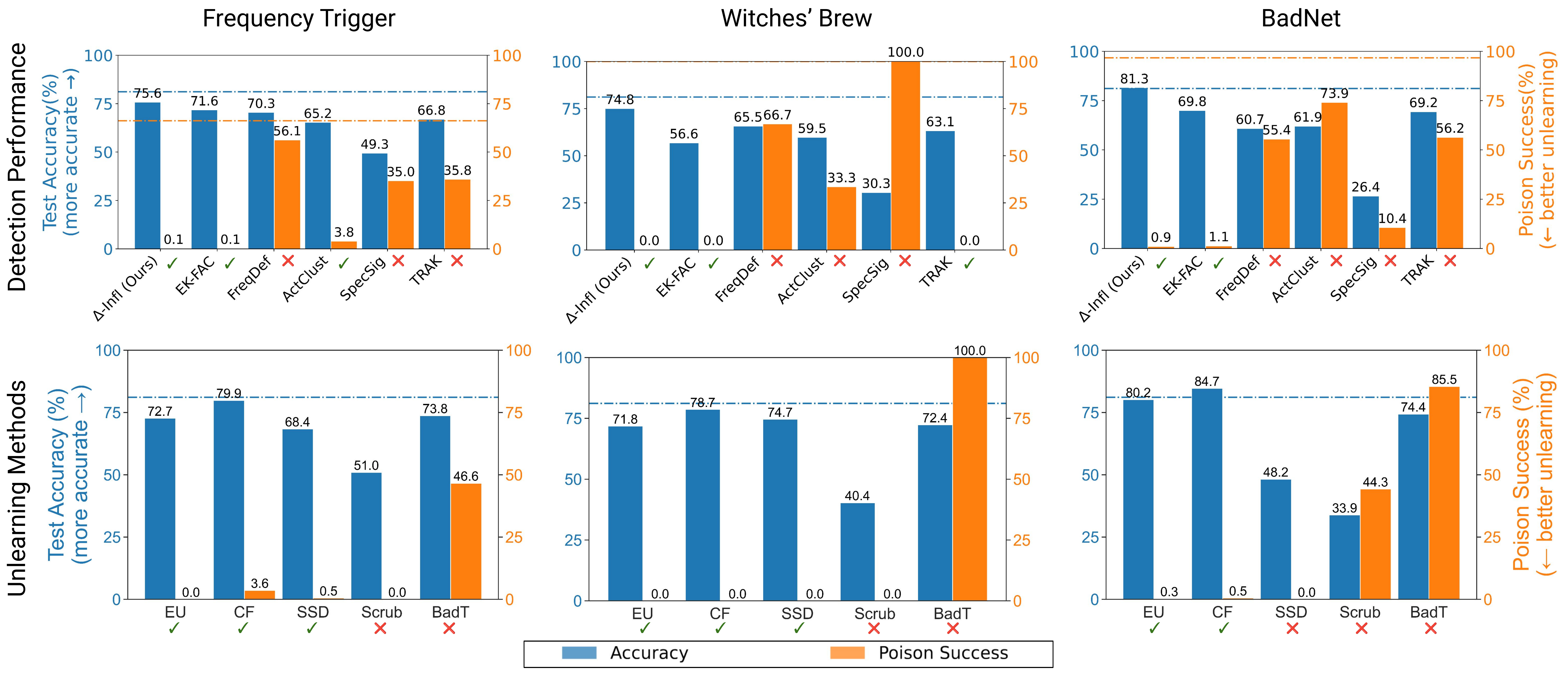}
    \caption{\textbf{Scaling to Imagenette.} In the top row, results on Imagenette are consistent with previous findings: \delinfl effectively unlearns all three types of poisons while preserving high test accuracy. In contrast, other detection methods often fail to unlearn or do so at the expense of test accuracy. In the bottom row, EU and CF consistently perform well.}
% with CF achieving the smallest accuracy drop and only a minor increase in poison success rate.}
    \label{fig:main-imagenette}
\end{figure*}

\subsection{The Causal Pitfall of Targeting a Poisoned Training Sample}
\label{sec:training_or_test_target}

\noindent\textcolor{softred}{\textbf{Setup.}} We investigate whether directly using a known poisoned training sample as the attribution target for influence functions can still effectively detect and unlearn poisons (although such availability can be hard to achieve for attacks like Witches' Brew).

\noindent\textcolor{softred}{\textbf{Results.}} As shown in \Cref{tab:knowing_train}, 
% using the poisoned training sample as target does not guarantee successful unlearning.
for CIFAR-10, when the attribution target is a training poison, \delinfl fails to achieve successful unlearning. This is indicated by the Poison Success Rate remaining at 100\%, which signifies that the attack remains fully effective. 
This inconsistent performance demonstrates that poisoned training samples are unreliable attribution targets for influence-based detection.

\begin{table}[h]
\centering
\caption{\textbf{Failed Unlearning When Targeting a Known Poisoned Training Point.} Comparison of using an affected Test point versus a known Train poison as the attribution target.}
\vspace{0.5em}
\scriptsize
\resizebox{0.7\linewidth}{!}{
\begin{tabular}{cccc}
\toprule
\textbf{Identified Point} & \textbf{TPR}($\uparrow$) & \textbf{Poison Success Rate} ($\downarrow$) & \textbf{Test Accuracy} ($\uparrow$) \\
\midrule
\multicolumn{4}{c}{\textbf{CIFAR10}} \\
\midrule
Test      & 19.4\%  & \textcolor{softgreen}{0\%}    & 91.0\%  \\
Train  & 8.4\% & \textcolor{softred}{100\%}  & 90.3\%  \\
\midrule
\multicolumn{4}{c}{\textbf{CIFAR100}} \\
\midrule
Test        & 62.4\% & \textcolor{softgreen}{0\%}    & 71.9\%  \\
Train  & 84.0\%  & \textcolor{softgreen}{0\%}  & 71.9\%  \\
\bottomrule
\end{tabular}}
% \caption{\textbf{Can we trust a poison from the training set?} We find that, for Witches' Brew, even when we are able to reveal a poisoned training sample, calculating influence scores on it cannot guarantee the model's recovery.}
\label{tab:knowing_train}
\end{table}

\noindent\textcolor{softred}{\textbf{Conclusion.}} These results justify our choice of using the attribution target from the deployment phase (\textit{i.e.}, test point) instead of training phase. The latter approach yields inconsistent unlearning performance because the causal dependency among poison peers is weaker than the collective influence of all poisons on an affected test point. Our choice is therefore grounded in a causal perspective: the goal is to find training examples responsible for a specific erroneous prediction, making the prediction itself the logical starting point. See Appendix~\ref{sec:appendix_dont_use_poisoned_train_sample} for a more detailed discussion.

\section{Scaling Findings to ImageNette}

\noindent\textcolor{softred}{\textbf{Setup.}} To evaluate the scalability and consistency of our \delinfl~algorithm
on a more complex and larger dataset, we conduct experiments on
Imagenette. The setup is consistent with the experiments
in~\Cref{sec:exp} with specific adjustments to accommodate
Imagenette's larger image sizes and increased complexity.
Specifically, we increase the patch size for BadNet poisoning, use a
larger trigger pattern for frequency-based poisoning, and poison a
greater fraction of training images (10\%). Additionally, for the
Witches' Brew method, we relax the perturbation constraint, setting
\(\epsilon=32\) instead of \(\epsilon=16\). \vspace{0.1cm}

\noindent\textcolor{softred}{\textbf{Results.}} Replicating our prior experiments on Imagenette,
\Cref{fig:main-imagenette} illustrates that \delinfl~continues to achieve
the most effective poison unlearning across all attack types,
maintaining minimal accuracy loss.  Notably, the EK-FAC baseline also
successfully unlearns all poisons but incurs a higher false positive
rate, leading to significant drops in test accuracy due to the
unnecessary removal of clean samples.  Additionally, when applying
various unlearning algorithms to the samples identified by \delinfl,
both CF and EU perform
consistently well with CF achieving notably higher accuracy during
poison unlearning compared to EU.\vspace{0.1cm}

\noindent\textcolor{softred}{\textbf{Conclusion.}} Scaling to larger datasets preserves all prior conclusions, underscoring the robustness of our results and the effectiveness of \delinfl and CF-based unlearning methods.

\subsection{Scaling to Larger Set of Identified Poisoned Test Samples}

\begin{table}[h]
\centering
\resizebox{0.75\linewidth}{!}{
\begin{tabular}{ccccc}
\toprule
\textbf{Influence Methods} & \textbf{Precision}($\uparrow$) & \textbf{TPR}($\uparrow$) & \textbf{Poison Success Rate} ($\downarrow$) & \textbf{Test Accuracy} ($\uparrow$) \\
\midrule
\multicolumn{5}{c}{\textbf{1 identified test point}} \\
\midrule
EK-FAC        & 22.1\% & 99.1\%  & \textcolor{softgreen}{0.3\%}    & 68.7\%  \\
\(\Delta\)-Influence  & 49.0\% & 100\% & \textcolor{softgreen}{0.8\%}  & 79.7\%  \\
\midrule
\multicolumn{5}{c}{\textbf{5 identified test points}} \\
\midrule
EK-FAC         & 25.9\% & 98.8\% & \textcolor{softgreen}{0.5\%}    & 73.3\%  \\
EK-FAC(boosted)        & 34.2\% & 98.5\% & \textcolor{softgreen}{0.8\%}    & 75.4\%  \\
\(\Delta\)-Influence  & 66.7\% & 100\%  & \textcolor{softgreen}{0.5\%}  & 80.0\%  \\
\midrule
\multicolumn{5}{c}{\textbf{10 identified test points}} \\
\midrule
EK-FAC         & 26.6\% & 98.8\% & \textcolor{softgreen}{0.5\%}    & 75.8\%  \\
EK-FAC(boosted)        & 48.9\% & 97.2\%  & \textcolor{softgreen}{1.6\%}    & 77.8\%  \\
\(\Delta\)-Influence  & 67.2\% & 100\%  & \textcolor{softgreen}{0.8\%}  & 79.9\%  \\
\bottomrule
\end{tabular}}
\vspace{0.2cm}
\caption{\textbf{ImageNette BadNet.} For BadNet poison on the ImageNette dataset, increasing the number of identified test points significantly improves the precision. This enhancement leads to a notable reduction in false positives, thereby achieving higher overall test accuracy.}
\vspace{-0.4cm}
\label{tab:more_points_badnet}
\end{table}

\begin{table}[h]
\centering
\resizebox{0.75\linewidth}{!}{
\begin{tabular}{ccccc}
\toprule
\textbf{Influence Methods} & \textbf{Precision}($\uparrow$) & \textbf{TPR}($\uparrow$) & \textbf{Poison Success Rate} ($\downarrow$) & \textbf{Test Accuracy} ($\uparrow$) \\
\midrule
\multicolumn{5}{c}{\textbf{1 identified test point}} \\
\midrule
EK-FAC        & 10.5\% & 99.3\%  & \textcolor{softgreen}{0\%}    & 72.4\%  \\
\(\Delta\)-Influence  & 25.8\% & 99.3\% & \textcolor{softgreen}{0\%}  & 75.4\%  \\
\midrule
\multicolumn{5}{c}{\textbf{5 identified test points}} \\
\midrule
EK-FAC         & 12.8\% & 99.0\% & \textcolor{softgreen}{0\%}    & 74.4\%  \\
EK-FAC(boosted)        & 21.8\% & 99.0\% & \textcolor{softgreen}{0.3\%}    & 74.0\%  \\
\(\Delta\)-Influence  & 27.5\% & 99.3\%  & \textcolor{softgreen}{0.3\%}  & 76.6\%  \\
\midrule
\multicolumn{5}{c}{\textbf{10 identified test points}} \\
\midrule
EK-FAC         & 12.9\% & 99.3\% & \textcolor{softgreen}{0\%}    & 74.1\%  \\
EK-FAC(boosted)        & 24.2\% & 99.3\% & \textcolor{softgreen}{0.3\%}    & 73.6\%  \\
\(\Delta\)-Influence  & 28.7\% & 99.3\%  & \textcolor{softgreen}{0.3\%}  & 75.0\%  \\
\bottomrule
\end{tabular}}
\vspace{0.2cm}
\caption{\textbf{ImageNette Frequency Trigger.} For frequency trigger poison on the ImageNette dataset, increasing the number of identified test points significantly improves the precision. This enhancement leads to a notable reduction in false positives, thereby achieving higher overall test accuracy.}
\vspace{-0.2cm}
\label{tab:more_points_freq}
\end{table}

\noindent\textcolor{softred}{\textbf{Setup.}} For attack methods such as Witches' Brew, only a single affected test point is identified. However, in cases where multiple test points can be identified, such as with BadNet Patch and Smooth Trigger attacks, we explore ways to enhance performance using two influence-based methods: \delinfl and EK-FAC, on the ImageNette dataset. Specifically, we show results when selecting five and ten test points to identify corresponding input points and determine their intersection as the poisoned data across both methods. This is done similarly to the \delinfl algorithm by retaining points with influence higher than the tolerance threshold, hence EK-FAC is additionally labeled (boosted).

\noindent\textcolor{softred}{\textbf{Results.}} We showcase performance in  \Cref{tab:more_points_badnet} for BadNet poison and  \Cref{tab:more_points_freq} for frequency trigger poison respectively. We observe a consistent trend: as the set of identified poisons increases, the precision improves significantly, leading to a substantial reduction in false positives and ultimately higher test accuracy. Overall, identifying multiple poisoned test points enables more precise detection of poisons in the training set when using \delinfl-like aggregation algorithms across test poisoned points.

\vspace{-0.2cm}
\section{Conclusion}
\vspace{-0.15cm}
In this study, we address a critical issue in corrective machine unlearning: identifying key training samples whose removal can unlearn a data poisoning attack. We address a practical scenario where only a limited number of affected test points are known—potentially discovered post-deployment or through internal testing. To this end, we introduce \delinfl, a novel approach that uses influence functions to trace abnormal model behavior back to the responsible poisoned training data, requiring as little as one affected test example. By retraining without these identified points, $\Delta$-Influence successfully unlearns multiple poisoning attacks across diverse datasets. We evaluate our method against five state-of-the-art detection algorithms and apply five well known unlearning algorithms to the identified training set. Our results demonstrate that $\Delta$-Influence consistently outperforms existing approaches in all tested scenarios. Our findings highlight the potential of influence functions as a foundation for unlearning data poisoning attacks. Additionally, our ablation study sheds light on the strengths and limitations of various poisoning attacks and unlearning algorithms, offering insights that could inform the development of more effective unlearning techniques and robust poisoning attacks for rigorous testing.

\section*{Acknowledgements}

The authors would like to thank (in alphabetic order): Shashwat Goel, Shyamgopal Karthik, Elisa Nguyen, Shiven Sinha, Shashwat Singh,  Matthias Tangemann, Vishaal Udandarao for their helpful feedback. WL, JL, and CSW acknowledges support from the Supervised Program for Alignment Research (SPAR) research program. We also acknowledge the Center for AI Safety (CAIS) for their support in providing the computational resources necessary for this study.

%AP acknowledges financial support via the Open Philantropy Foundation funded by the Good Ventures Foundation. 
{
    \small
    \bibliographystyle{plainnat}
    \bibliography{bibliography}
}

%%%%%%%%%%%%%%%%%%%%%%%%%%%%%%%%%%%%%%%%%%%%%%%%%%%%%%%%%%%%
\clearpage
\appendix
 %\maketitlesupplementary
% { {\section*{\centering \LARGE Appendix}}}\\
\section{Connections to Existing Work}
\label{sec:appendix_existing_work}

\noindent\textbf{Data Attribution: A Brief Overview}

The problem of training data attribution (TDA) has been explored using various approaches such as influence functions~\citep{koh2017understanding,koh2019accuracy}, Shapley value-based estimators~\citep{ghorbani2019data}, empirical influence computation~\citep{feldman2020neural}, and predictive datamodels~\citep{park_2023_trak}. 

Broadly, TDA methods can be categorized into three groups: retraining-based methods, gradient-based methods, and predictive attribution models~(see~\cite{hammoudeh2024training} for a survey). Retraining-based methods systematically retrain models with and without specific training samples and observe changes in the model’s outputs~\citep{ghorbani2019data,jia2019towards,feldman2020neural}. While these methods yield relatively accurate influence scores, they are computationally prohibitive for moderately large models, as the number of retrains often grows with the size of the training data. Gradient-based methods, such as influence functions~\citep{cook1980characterizations}, are computationally cheaper but often produce less reliable influence estimates for complex models~\citep{basu2020influence}.

Influence functions approximate the effect of individual training samples on a model's predictions by measuring how a prediction changes when a sample's weight is slightly perturbed. They were introduced to machine learning by~\citet{koh_understanding_2017} and have since been refined~\citep{grosse_studying_2023,kim2024gex,pruthi2020estimating}. In data poisoning contexts, \citet{seetharaman_influence_2022} used influence functions to mitigate degradation caused by previously identified poisoned data~\citep{steinhardt_certified_2017}. Building on this, we explore how advanced influence functions like EK-FAC~\citep{grosse_studying_2023} can identify training examples disproportionately contributing to anomalous predictions in poisoned models.

Another approach, predictive data attribution, focuses on predicting model behavior directly based on training data~\citep{ilyas2022datamodels,park_2023_trak}. While this approach can provide accurate influence estimates, the cost of training predictive models remains a significant limitation.

\noindent\textbf{Unlearning: A Brief Overview}

Machine unlearning, first proposed by~\citet{Cao2015Unlearning}, enables ML models to ``forget" specific data points by removing their influence. This concept has gained importance with data protection regulations such as GDPR in the EU, which enforce the "right to be forgotten." Ideally, unlearning produces models equivalent to retraining from scratch after excluding the target data~\citep{Cao2015Unlearning,bourtoule_machine_2021,gupta2021adaptive}. However, retraining is computationally expensive, leading to the development of approximate unlearning methods~\citep{ginart2019deletion,guo2020certified,neel2020descenttodelete}. These methods are often inspired by concepts from differential privacy, with the relevant~(\((\epsilon,\delta\))-provable unlearning definition formalized in~\citet{sekhari2021remember}.

Recently, the scope of machine unlearning has expanded beyond privacy to address post-hoc system degradation, such as harmful knowledge removal~\citep{li2024wmdp} and adversarial attacks~\citep{pawelczyk_machine_2024,goel_corrective_2024,schoepf2024potion}. In corrective unlearning, \citet{pawelczyk_machine_2024} demonstrated the difficulty of mitigating strong poisons like Witches' Brew, while~\citet{goel_corrective_2024} highlighted challenges when the complete set of manipulated data is unknown. These complexities underscore the inherent difficulty of the setting we address in this work.

\noindent\textbf{Data Poisoning Attacks} 
Data poisoning attacks are a significant threat to ML systems due to their ease of deployment and difficulty in detection. Even minor modifications to training data can lead to successful attacks on models trained on large datasets~\citep{carlini2023poisoning}. In this paper, we consider three forms of data poisoning attacks: a \textit{backdoor attack}~\citep{gu_2017_badnets} that adds a small patch in the corner of attacked images and modifies their labels to a target label, a \textit{smooth trigger attack}~\citep{zeng_rethinking_2021} that adds a trained pattern which is both hard to identify either in raw image domain or frequency domain, and \textit{Witches' Brew}~\citep{geiping_witches_2021}, which adds a trained imperceptible pattern on attacked images without modifying labels. Note that the first two attacks modify the victim images' labels, while Witches' Brew is a \textit{clean-label attack}.\vspace{0.1cm}

\noindent\textbf{Data Poisoning Defences}
Defenses against data poisoning often involve trigger-pattern reverse engineering using clean data~\citep{wang_neural_2019,guo_tabor_2019,tao_better_2022,dong_black-box_2021,wang_practical_2020}. These methods require additional steps such as input pre-filtering, neuron pruning, or fine-tuning~\citep{liu_fine-pruning_2018,chen2019deepinspect,li_neural_2021,zeng_adversarial_2022}. Other approaches, like Anti-Backdoor Learning~\citep{li_anti-backdoor_2021} and BaDLoss~\citep{alex_2024_protecting}, necessitate tracking model updates and clean training samples, adding complexity to the defense process.

In contrast, our method requires access only to the trained model and a single poisoned test example, offering a simpler yet effective defense mechanism.

% \section{Augmentations Used in \(\Delta\)-Influence}
% \clearpage
\section{Experiment Details}
\label{sec:appendix_exp_details}

\subsection{Poisoned Training Sample Is Not a Reliable Target for Influence-Based Unlearning}
\label{sec:appendix_dont_use_poisoned_train_sample}

\hspace*{\parindent}Given that a small subset of poisoned training data—commonly referred to as a forget set~\citep{goel_corrective_2024}, could be identified as the prerequisite for unlearning, 
a natural question arises: why \delinfl focus on an identified affected test sample rather than simply using a poisoned training sample?

One overlooked fact is for some covert attack like Witches' Brew, the attack pattern is different between training and testing, which is not the case for Frequency Trigger and BadNet. 
Moreover, the clean-label attack manner and the imperceptible perturbations make it notoriously difficult to identify training poisons for such attacks. 
However, we emphasize that, regardless of how clever and stealthy an attack is designed, its primary goal is to alter model predictions on specific test points, making anomalies more apparent after deployment. Hence we think that having one identified test point is generally more feasible than identifying a poisoned training point in this context and better suited for influence-based analysis, which attributes model behavior to particular training instances.

What's more, as demonstrated in \Cref{tab:knowing_train}, taking Witches' Brew as an example, we find that even when defenders can reveal a poisoned training sample, the poisoned behavior cannot be reliably mitigated, while we show that \delinfl, utilizing an identified poisoned test point, can systematically undo the attack's impact.

This underscores a fundamental limitation of how influence functions work: influence functions inherently rely on clear causal relationships, where specific training samples directly impact corresponding test-time anomalies. However, in poisoned learning scenarios, such causality is often obscured: 
while the training poisons as a whole shifts model behavior, it's causal effect with one individual poisoned sample in it could be more ambiguous.
This intuition that using a poisoned training sample as the target is less reliable than using an affected test point, is further supported by our empirical findings 
(as shown in \Cref{tab:knowing_train}, influence-based methods fail to unlearn poisons when guided by a poisoned train point).
We hope this observation provides useful insights for how target selection impacts causal tracing effectiveness in influence-based unlearning.

Based on the above observations, we suggest that using an identified poisoned test point for influence-based unlearning. 
Although when the attack pattern is consistent between training and testing (\textit{e.g.} Frequency Trigger~\citep{zeng_rethinking_2021} and BadNet~\citep{gu_2017_badnets}), using a poisoned training sample as the target also work, 
% we believe that as the defender, we should not assume we know what attack is performed, either have knowledge of the attack pattern. 
We argue that defenders should not assume such prior knowledge, \textit{e.g.} what attack is performed and what the attack pattern is, which is rarely available in practice. 
Hence using a poisoned test point is more reliable and generalizable across different attack scenarios.

Finally, although it's not the focus of this paper, here we discuss approaches to get such a test point, realistic scenarios include: (i) whitehat adversarial research teams conducting jailbreaking-style tests to expose failure modes; (ii) Companies internally systematically stress-testing for vulnerabilities; and (iii) Companies using anomaly detection algorithms to monitor user interactions for abnormal behavior. Note that determining whether a test point is harmful or benign relies on the developer's domain expertise, this largely unexplored area is increasingly necessary due to massive training datasets and the rise of opaque open-source base models, offering promising directions for future research. 

\subsection{Predefined Set for Image Augmentations}
\label{sec:predefined_transformations}
We employ a predefined set of standard image augmentation techniques: Flip, Rotation, Color Jitter, Elastic Transformation, Blur, Inversion, Color Switch, and Random Affine transform. For each transform, one augmentation is randomly selected from this set and applied to the affected test image.

% \subsection{The Predefined Set for Image Augmentations}
% We utilize a set of standard image augmentations: Flip, Rotation, Color Jitter, Elastic, Blurrer, Inverter, Color Switch and Random Affine. For each transformation \(g_j\), one augmentation will be randomly picked from this set and applied onto the affected test image.

\vspace{-0.15cm}
\subsection{Attack Methods}
\vspace{-0.1cm}
The attack target and victim class are chosen at random for each trial. We shall now discuss the details for each attack method below. The relevant code is additionally publicly available in our repository: \url{https://github.com/Ruby-a07/delta-influence}.\vspace{0.1cm}

\noindent\textbf{BadNet} For CIFAR datasets, we add a $3 \times 3$ checkboard-patterned black patch (pixel values set to zero) at the bottom-right corner of each $32 \times 32$ image. For the Imagenette dataset, we utilize a larger square $22 \times 22$ black patch to ensure successful injection of the poison. The number of poisoned images varies by dataset: 500 for CIFAR10, 350 for CIFAR100, and 858 for Imagenette.\vspace{0.1cm}

\noindent\textbf{Smooth Trigger} The smooth trigger is generated for each dataset following the algorithm proposed in~\citep{zeng_rethinking_2021}.  The number of poisoned images similarly varies by dataset: 500 for CIFAR10, 125 for CIFAR100 and 300 for Imagenette. Since the poison is more powerful, we are able to poison the model with less number of poisoned samples.\vspace{0.1cm}

\noindent\textbf{Witches' Brew} The adversarial pattern is generated according to the method described in~\citep{geiping_witches_2021}. The number of poisoned images similarly varies by dataset: 500 for CIFAR10, 125 for CIFAR100 and 947 for ImageNette respectively. To ensure successful poisoning of Imagenette, we set we set eps=32, which is twice the value used for CIFAR10 and CIFAR100 (eps=16).

\vspace{-0.15cm}
\subsection{Hyperparameters for Detection Methods}
\label{sec:appendix_detection_hparams}
\vspace{-0.1cm}

The hyperparameters are optimized through a grid search process to find the best possible values, following the process from \citet{goel_corrective_2024}. Specifically:\vspace{0.1cm}

\noindent\textbf{ActClust}
We set the number of components,  \(n_{comp}=3\), for all experiments. ActClust is quite robust a method, and we find that a value of 3 performs consistently best across all experiments.\vspace{0.1cm}

\noindent\textbf{SpecSig}
SpecSig involves two hyperparameters: the spectral threshold, used to identify significant singular values, and the contribution threshold, used to identify significant data point contributions. SpecSig is sensitive to both parameters. Typically, we select the best spectral threshold by grid search per dataset from the values {4, 6, 8, 10} and the contribution threshold from {7, 9, 11, 13}. Higher values indicate a stricter constraint, resulting in fewer detected examples.\vspace{0.1cm}

\noindent\textbf{FreqDef}
For datasets with different image sizes, we train a specialized classifier following the methodology described in~\citep{zeng_rethinking_2021}.\vspace{0.1cm}

\noindent\textbf{EK-FAC}
We typically begin with a threshold value of 0 and select the best threshold among values (0, 10, 100, 500). Higher threshold values imply stricter filtering constraints, leading to fewer detected examples.\vspace{0.1cm}

\noindent\textbf{Ours} Similar to EK-FAC, starting with a threshold \(\tau\) of 0 is generally effective where we search over (0, -1, -5, -10, -100). Lower threshold values and smaller \(n_{\mathrm{tol}}\) indicate stricter filtering constraints. For \(n_{\mathrm{tol}}\), We normally search over {0, 1, 2, 3}, with 1 proving to be effective in most cases. 

\vspace{-0.15cm}
\subsection{Hyperparameters for SSD}
\vspace{-0.1cm}

Among the five unlearning methods considered, SSD is particularly sensitive to hyperparameters but is computationally efficient. This allows for lots of runs to select the optimal unlearning result. For each experiment, we evaluate all possible combinations of two SSD hyperparameters, the weight selection threshold, which controls how protective the selection should be, and the weight dampening constant which defines the level of parameters protection. Specifically, we choose the weight selection threshold from values {2, 10, 50} and the  weight dampening constant from {0.01, 0.1, 1}.

% \clearpage

\section{Results for Ablating Image-Only and Label-Only Augmentations}

\begin{figure}[h]
    \centering
    \includegraphics[width=\linewidth]{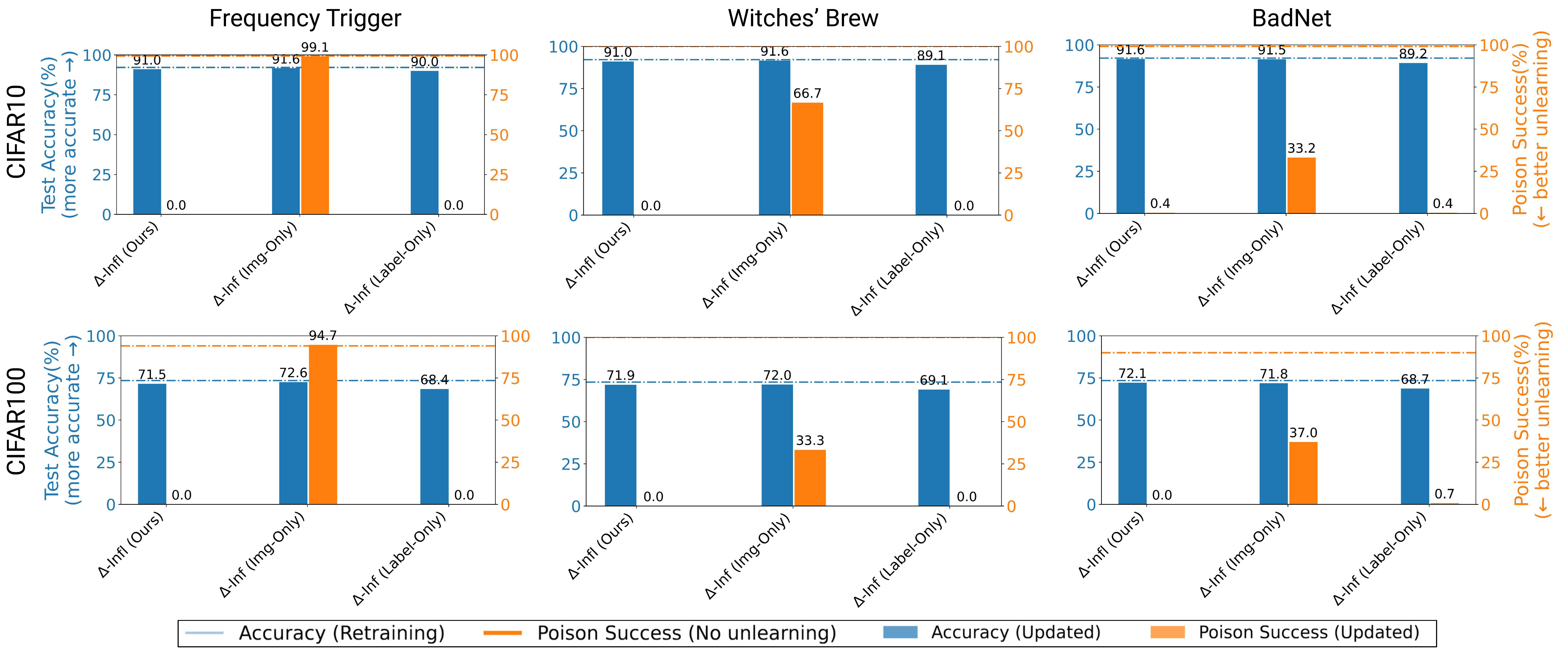}
    \caption{\textcolor{kleinblue2}{\underline{Poison Success Rate}} and \textcolor{kleinblue2}{\underline{Test Accuracy.}} This table shows both poison unlearning effectiveness and model utility. A method is considered successful if the poison success rate is below 5\%. Label augmentations are instrumental towards identifying poisons, even in the clean-label poison cases. Figure structure from \citep{pawelczyk_machine_2024}.}
    \label{fig:abl_unlearn}
\end{figure}

\label{sec:appendix_additional_exps}

We show in~\Cref{fig:abl_unlearn} that Label-Only augmentations are effective in removing the data poisoning (lower poison success rate), while Image-Only augmentations perform poorly in this regard. However, as demonstrated in \Cref{tab:combined_3and4}, Label-Only augmentations lead to the unnecessary discard of many clean samples, whereas image augmentations significantly reduce the false positive rate, preserving clean data and improving detection precision. Therefore both label and image augmentations are crucial to the effectiveness of the $\Delta$-Influence method.

\section{Does $\Delta$-Influence Perform the Best Across Unlearning Algorithms?}

\noindent\textcolor{softred}{\textbf{Setup.}} The probe was conducted across various detection methods; however, instead of employing the exact unlearning algorithm, we use a popular alternative algorithm called SCRUB which involves gradient ascent. We similarly measure the performance as well as the success rate of the poison removal were evaluated. Note the TPR rate and precision do not change.

\begin{figure}[t]
    \centering
   
    \includegraphics[width=\linewidth]{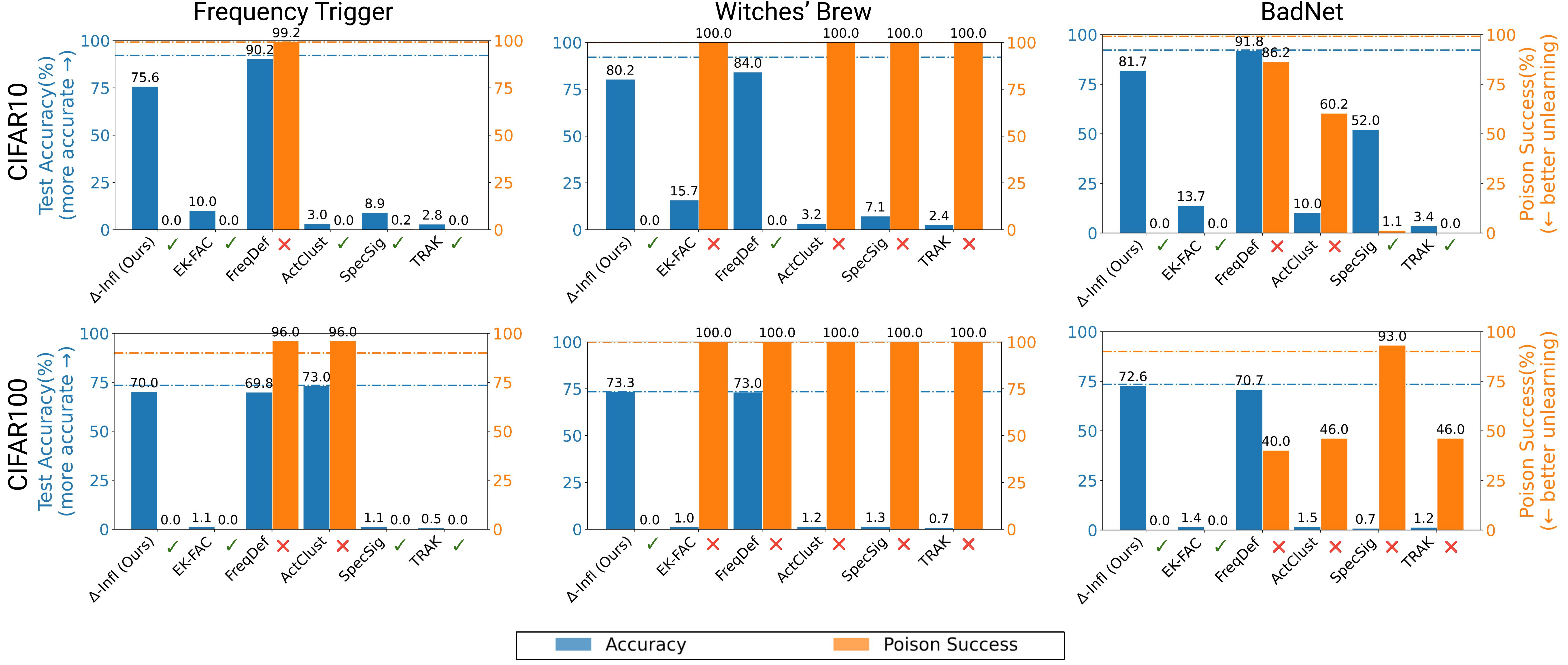}
    \caption{\textcolor{kleinblue2}{Poison Success Rate} and \textcolor{kleinblue2}{Test Accuracy. with SCRUB Unlearning algorithm.} This table shows both poison unlearning effectiveness and model utility. A method is considered successful if the poison success rate is below 5\%, marked by \textcolor{softgreen}{\checkmark}, with unsuccessful methods marked by \textcolor{softred}{\textbf{$\times$}}. $\Delta$-Influence is successful in 6/6 cases, while the rest fail by not be distinguishable from a randomly initialized model. In contrast, $\Delta$-Influence has only minor drops in test accuracy. Figure structure from \citep{pawelczyk_machine_2024}.}
    \label{fig:scrub_unlearn}
\end{figure}

\noindent\textcolor{softred}{\textbf{Results.}} The evaluation results in \Cref{fig:scrub_unlearn} shows that \delinfl outperforms other methods, unlearning poisons in all six cases with minimal performance loss. In contrast, EK-FAC, ActClust, and SpecSig performed randomly,  achieving unlearning primarily because even a randomly initialized model would not retain poisoning. Performance drops were primarily due to SCRUB's sensitivity to false positives from its gradient ascent step. FreqDef avoided randomness but failed to unlearn poisons in all cases. Notably, \delinfl minimized false positives, maintaining consistent and reliable outcomes.

\noindent\textcolor{softred}{\textbf{Conclusions.}} \delinfl proves to be remarkably robust even across unlearning methods which are highly sensitive to false positives. It achieves a 6/6 poison removal rate while incurring only minor performance losses due to false positives.

\section{Limitations}
\label{sec:limitations}

\delinfl is based on influence functions and hence inherit their drawbacks. Possible attacks like those are only injected during test phase can evade our detection.

\section{Compute Resources}

We used a single NVIDIA A100-80GB GPU for all experiments. The training time for each experiment varied depending on the model, dataset, and hyperparameters. On ImageNette poisoned by Witches' Brew (with the according hyperparameters provided in the appendix), the time of detection is normally half an hour. 
Note that this could be speed up by parallelizing the computation of influence scores but this is not the focus of this paper.

\end{document}